\documentclass[letterpaper]{article} 
\usepackage{aaai25}  
\usepackage{times}  
\usepackage{helvet}  
\usepackage{courier}  
\usepackage[hyphens]{url}  
\usepackage{graphicx} 
\usepackage{natbib}  
\usepackage{caption} 
\usepackage{algorithm}
\usepackage{algorithmic}

\usepackage{newfloat}
\usepackage{listings}
\DeclareCaptionStyle{ruled}{labelfont=normalfont,labelsep=colon,strut=off} 
\floatstyle{ruled}
\newfloat{listing}{tb}{lst}{}
\floatname{listing}{Listing}
\newcommand{\system}{AMC}
\usepackage{threeparttable}
\usepackage{makecell}
\usepackage{booktabs}
\usepackage{multirow}
\usepackage{bbding}
\usepackage{rotating}
\usepackage{tabularx}
\usepackage{amsmath}
\usepackage{cleveref}
\usepackage{pifont}
\usepackage[table]{xcolor}

\nocopyright

\newcommand{\figref}[1]{Figure \ref{#1}}

\title{Depression Diagnosis Dialogue Simulation: Self-improving Psychiatrist with Tertiary Memory}

\author {
    Kunyao Lan\textsuperscript{\rm 1\dag},
    Bingrui Jin\textsuperscript{\rm 1\dag},
    Zichen Zhu\textsuperscript{\rm 1},
    Siyuan Chen\textsuperscript{\rm 1},\\
    Shu Zhang\textsuperscript{\rm 2},
    Kenny Q. Zhu\textsuperscript{\rm 3},
    Mengyue Wu\textsuperscript{\rm 1\ddag}
}
\affiliations {
    \textsuperscript{\rm 1}X-LANCE Lab, Dept. of Computer Science and Engineering\\
    MoE Key Lab of Artificial Intelligence, AI Institute\\
    Shanghai Jiao Tong University, Shanghai, China\\
    \textsuperscript{\rm 2}Shanda Group, Shanghai, China\\
    \textsuperscript{\rm 3}University of Texas at Arlington, Arlington, Texas, USA\\
    \texttt{\{lankunyao,suzy\_jin,jameszhuthethird,chensiyuan925,mengyuewu\}@sjtu.edu.cn, zhangshu@shanda.com, kenny.zhu@uta.edu}
}

\begin{document}

\maketitle
\renewcommand{\thefootnote}{\fnsymbol{footnote}}
\footnotetext[2]{This work was partially done during an internship at Shanda Group.}
\footnotetext[3]{Corresponding authors.}

\begin{abstract}
Mental health issues, particularly depressive disorders, present significant challenges in contemporary society, necessitating the development of effective automated diagnostic methods. This paper introduces the Agent Mental Clinic (AMC), a self-improving conversational agent system designed to enhance depression diagnosis through simulated dialogues between patient and psychiatrist agents. To enhance the dialogue quality and diagnosis accuracy, we design a psychiatrist agent consisting of a tertiary memory structure, a dialogue control and reflect plugin that acts as ``supervisor'' and a memory sampling module, fully leveraging the skills reflected by the psychiatrist agent, achieving great accuracy on depression risk and suicide risk diagnosis via conversation. Experiment results on datasets collected in real-life scenarios demonstrate that the system, simulating the procedure of training psychiatrists, can be a promising optimization method for aligning LLMs with real-life distribution in specific domains without modifying the weights of LLMs, even when only a few representative labeled cases are available.
\end{abstract}

\section{Introduction}

Mental health issues represent a significant challenge in contemporary society. According to the World Health Organization (WHO)~\cite{Freeman2022TheWM}, approximately 298 million individuals are affected by depressive disorders, which are the most prevalent issue of mental disorders among adults. Concurrently, many mental health systems are characterized by substantial deficiencies and imbalances in resources and services~\cite{Freeman2022TheWM}. This situation highlights the urgent need for more effective automated methods for the detection and diagnosis of depression, which would lead to better allocation of medical resources.

Among various automated diagnostic approaches, conversational agents (CAs) have emerged as particularly effective due to their cost efficiency, time-saving capabilities, and ability to maintain user anonymity~\cite{Ma2023UnderstandingTB}. CAs are especially valuable in the diagnosis of mental illnesses, where the process is more complex than diagnosing physical conditions, which can often be confirmed through test reports. Mental health diagnoses rely heavily on qualitative information obtained through patient interactions, which presents unique challenges compared to the more straightforward use of structured electronic medical records (EMRs)~\cite{Li2024AgentHA}. Additionally, in contrast to self-assessment chatbots~\cite{Jaiswal2019VirtualHQ}, conversational agents powered by large language models (LLMs)~\cite{chen2023llmempoweredchatbotspsychiatristpatient} demonstrate a greater propensity to offer emotional support and apply advanced professional skills to enhance the accuracy and effectiveness of diagnostic tasks.

\begin{figure}[!ht]
    \centering
    \includegraphics[width=1\linewidth]{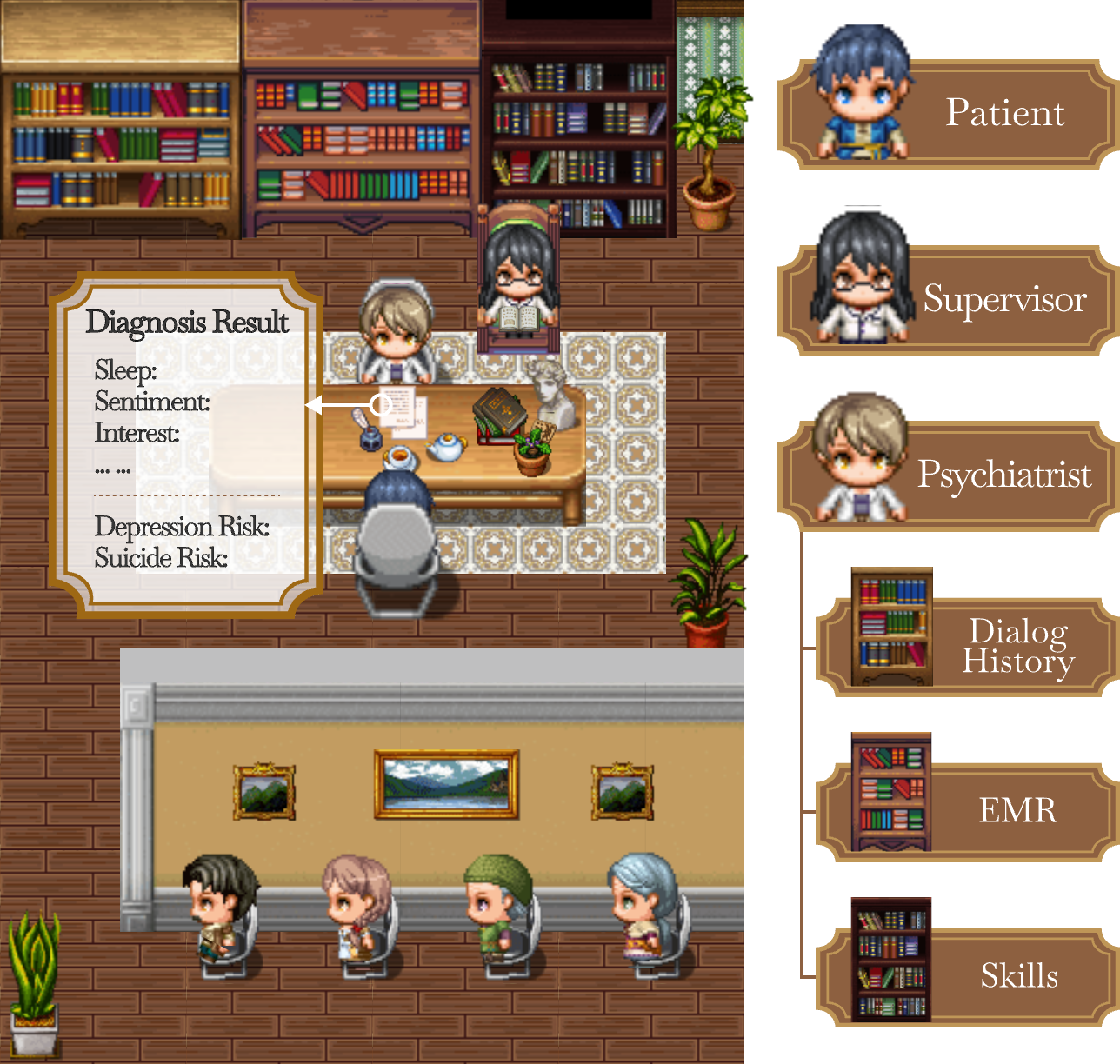}
    \caption{\textbf{Agent Mental Clinic (\system).} The symptom list is collected during the conversation between the psychiatrist agent and the patient agent, with the guidance of the supervisor. Dialog History, Electronic Medical Records (EMR), and Skills are hierarchical memory layers of the psychiatrist agent. These layers are progressively refined throughout the depression diagnosis session.}
    \label{fig:demo}
\end{figure}

\begin{figure*}[!ht]
    \centering
    \includegraphics[width=1\linewidth]{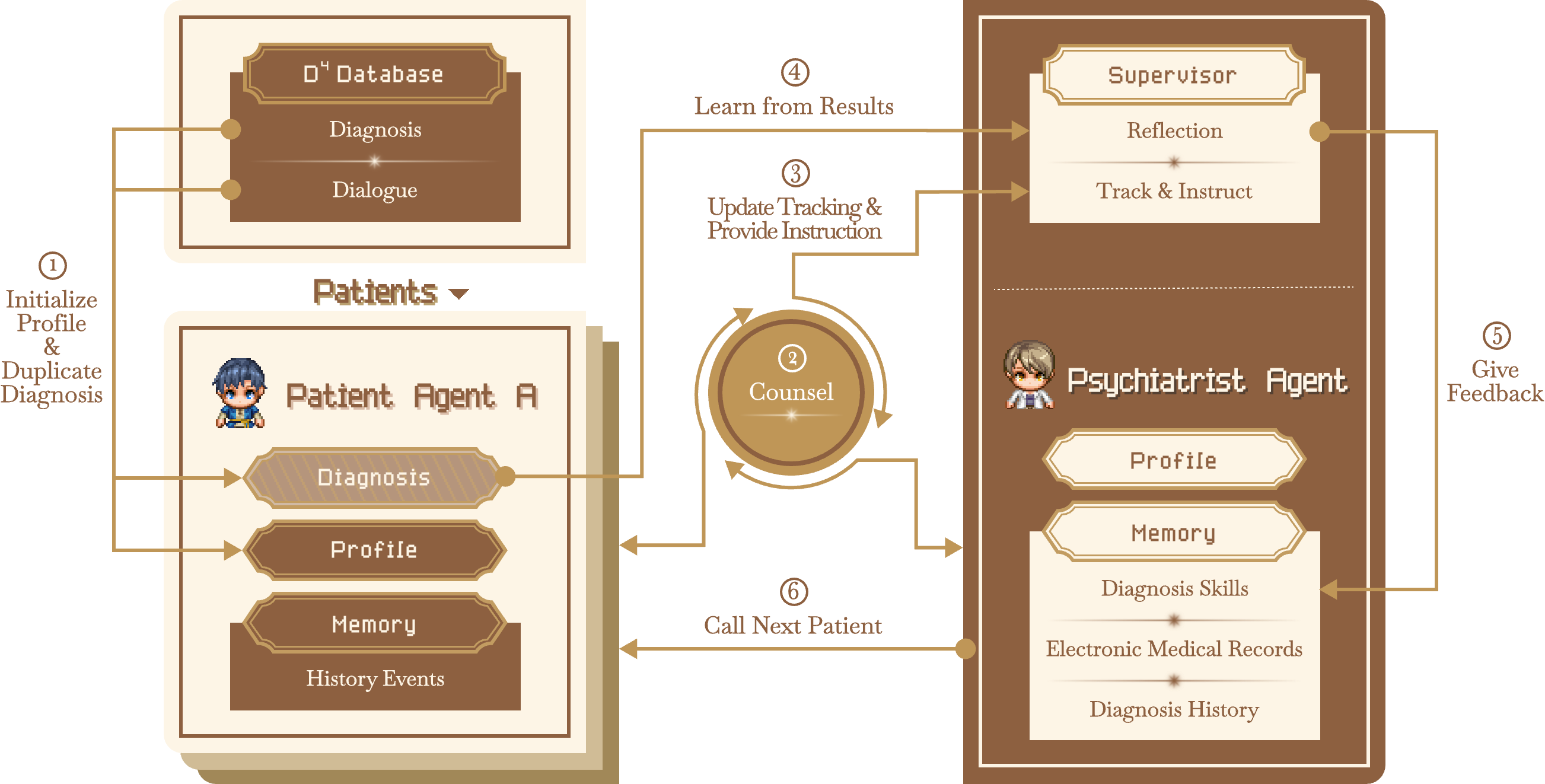}
    \caption{\textbf{The overview of \system.} Agents are divided into patient agents and psychiatrist agents, while the supervisor plugin is attached to the psychiatrist agents to control the dialogue process. D4 database is a dataset including patient portraits, doctor-patient dialogue, and patients' diagnostic summaries.}
    \label{fig:overview}
\end{figure*}

Despite the strong performance of LLM-empowered conversational agents (CAs) in various conversation-based tasks~\cite{wu2023autogenenablingnextgenllm, Shi2023LLMMiniCEXAE, chen2023llmempoweredchatbotspsychiatristpatient}, the inherent biases within LLMs~\cite{yeh-etal-2023-evaluating, xu2024prideprejudicellmamplifies} remain challenging, even with the use of advanced prompting techniques. To be specific, in the mental health domain, previous research has revealed that 1) LLMs exhibit a preference for certain strategies~\cite{kang2024largelanguagemodelsgood, chen2023llmempoweredchatbotspsychiatristpatient}, such as emotion and sleep-related symptoms. 2) In diagnosis sessions, LLMs prefer to ask all the questions in one turn rather than step by step, and come to a diagnosis conclusion in several turns.~\cite{chen2023llmempoweredchatbotspsychiatristpatient} 3) LLMs might overlook the potential mental states of patients~\cite{jin2024psyevalsuitementalhealth}. The bias of LLMs harms their effectiveness in tasks like providing emotional support and conducting diagnosis conversations.

These limitation of LLMs emphasizes the necessity of dynamic prompt modification and the development of novel generative agents~\cite{Park2023GenerativeAgents}. These agents can maintain long-term coherence in simulating human behavior by managing expansive memories. The design of memory structures~\cite{zhang2024surveymemorymechanismlarge} allows these agents to sustain coherence and stimulate their self-evolution~\cite{zhang2024training, zhang2024surveymemorymechanismlarge}, demonstrating effectiveness in reducing LLM biases and enhancing performance across various complex conversational tasks~\cite{qian2024chatdevcommunicativeagentssoftware,wu2023autogenenablingnextgenllm, Li2024AgentHA}.

Although the effectiveness of memory structure in agent dialogue simulation is widely acknowledged, most approaches of memory retrieval still rely on previous criteria~\cite{Park2023GenerativeAgents}, including relevance, recency, and importance, which proves inadequate in mental health diagnosis conversation since 1) GPT-generated importance scores often fail to align with the requirements of realistic depression diagnostic process. 2) Recency scores are not effective without time simulation in diagnosis simulation. 3) The prevalent Top-$k$ retrieval method is insufficient in increasing the diversity of memories accessed by agents, leading to redundancy as experiments progress. The redundancy not only increases the time-consuming of retrieval operations but also limits the potential for optimization.

To further explore the effectiveness of memory modules in agent dialogue simulation, we propose a novel conversational agents simulation system \textbf{A}gent \textbf{M}ental \textbf{C}linic (\system), focusing on depression diagnosis conversation sessions (\figref{fig:overview}). To replicate the diagnosis procedure in real life, we utilized D$^4$ dataset~\cite{yao-etal-2022-d4}, a Chinese depression diagnosis dialogue dataset based on user portraits collected in real-world settings to evaluate the depression diagnosis accuracy of the psychiatrist agents. The experimental results indicate that the psychiatrist agents can enhance their performance after reflecting and learning from training cases, and achieve a higher diagnosis accuracy on unknown cases.
The key contribution of this paper is as follows:

\begin{itemize}
    \item We introduce a novel conversational agent simulation system to imitate the diagnostic sessions between patient agents and psychiatrist agents, This system generates realistic depression diagnostic conversations between agents, providing a promising method for training intern psychiatrists and for conducting preliminary depression risk screening on individuals exhibiting depressive symptoms.
    \item We propose an innovative memory structure and a retrieval module to enhance the utilization of skills generated by the agents. This improvement suggests a direction for future optimization in both depression diagnosis and the simulation of diagnostic conversations.
    \item Experiment results based on depression diagnosis conversation collected from real-life scenarios demonstrate the effectiveness of our \system~system, with an average improvement of 6.05 percent in depression diagnosis, and 1.8 percent in suicide prediction. Our framework can be applied in other specific domains with only a limited number of finely labeled cases available for training.
\end{itemize}

\section{Mental Clinic Simulation}

The basic architecture of \system~follows the universal LLM-based conversational agent~\cite{Park2023GenerativeAgents} setting, as Figure~\ref{fig:overview} illustrated. The main characters defined in the \system~can be divided into three parts: 
\begin{itemize}
    \item A variety of patients, generated based on existing patient portraits in D$^4$;
    \item Psychiatrist, the agent that reflecting skills based on the diagnosis conversation;
    \item Supervisor, an incomplete agent without long-term memory, acts as an optional plugin of the psychiatrist, initialized at the beginning of each session, mainly controlling the dialogue process and stimulating the psychiatrist agents to reflect on the diagnosis results.
\end{itemize}

\paragraph{Patient Agents}

\begin{figure}[!ht]
    \centering
    \includegraphics[width=1\linewidth]{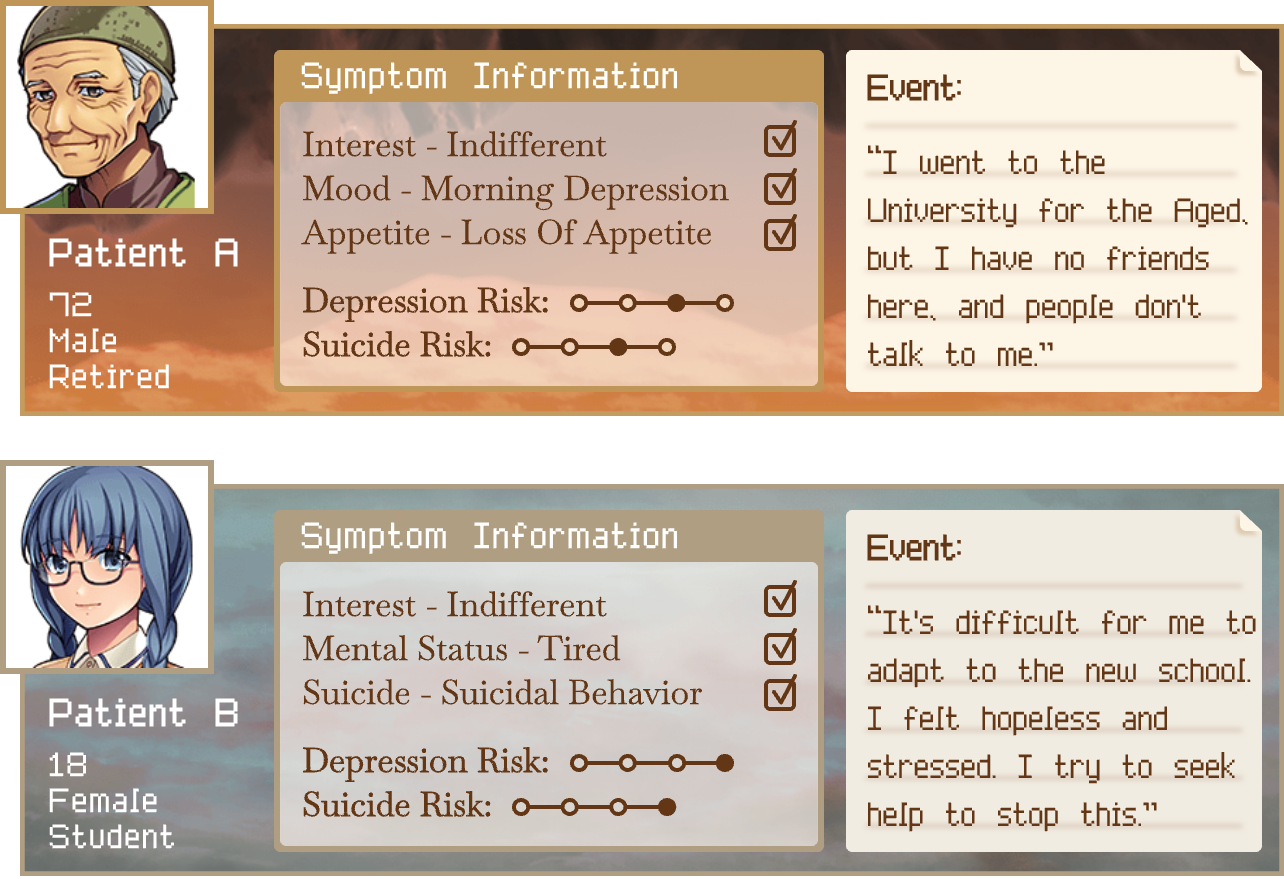}
    \caption{\textbf{Patient Profile Examples.}  We select two patient agents to illustrate. The profiles of patient agents are generated from cases of D$^4$.}
    \label{fig:portrait}
\end{figure}

Patient agents refer to agents playing individuals suffering from depressive moods and might be diagnosed with depressive disorder. They come to the clinic, seeking help from the psychiatrist agent to determine whether further intervention is necessary. To simulate depressive patients in a real-life scenario, we utilize a public quasi-clinical-standard depression diagnosis dialogue dataset D$^4$~\cite{yao-etal-2022-d4} to initialize patient agents, as step (1) in Figure \ref{fig:overview} represents. D$^4$ is a Chinese depression diagnosis dialogue dataset containing 1339 dialogues conducted by well-trained patients and psychiatrists. The data characteristics of the dataset are attached in the appendix. All the dialogues are collected based on portraits of real potential patients. We select 100 representative cases from the train set of D$^4$, two examples are shown in Figure~\ref{fig:portrait}. In the meantime, we also include the original 132 test cases in D$^4$ to conduct further experiments. Each selected case includes the patient profile, the dialogue history between the doctor and the patient, and the diagnosis result of the patient. \label{train/test}

To better simulate the symptom status of the patients, we adopt detailed symptom ontology defined in the Diagnostic and Statistical Manual of Mental Disorders (DSM-5-TR)~\cite{Arbanas2015DiagnosticAS} and prior work~\cite{Lan2024TowardsRA} to track the status of patients. However, patient agents are only provided with symptom lists, without the chief complaint and information about life events, resulting in poor role-play quality during the conversation, only replying whether they had the corresponding symptoms. Therefore, we use \texttt{GPT-4} to generate life event memory based on the dialogue history.

\paragraph{Psychiatrist Agent}
The psychiatrist agent is initialized with diagnosis skills from ICD-11\footnote{https://icd.who.int}. It leads a conversation with patient agents, collecting information to be noted in the symptom list, as step (2) in Figure~\ref{fig:overview} illustrates. At the end of the session, the psychiatrist agent summarizes the electronic medical record (EMR) and diagnoses the depression risk and suicide risk of the patient agent. After receiving the reflected skills provided by the supervisor plugin and storing them in the memory, the psychiatrist agent calls for the next patient and repeats the diagnosis session, as step (5) and step (6) in Figure~\ref{fig:overview} represent.

\paragraph{Supervisor Plugin}
The supervisor plugin acts as the coach or assistant of the psychiatrist agent, tracking the patient status and providing dialogue-controlling instructions during the diagnosis conversation, as step (3) in Figure \ref{fig:overview} shows. At the end of each session, the supervisor plugin compares the generated diagnosis results with the ground truth result diagnosed by a professional psychiatrist in D$^4$, reflects the skills that may be helpful to diagnose patients more precisely and provides them to the psychiatrist agent. We intentionally separate the supervisor as an individual role for future intervention from real psychiatrists, i.e. experts can intervene in the system to provide feedback and achieve a human-in-the-loop framework.

\section{Method}
\label{sec:method}
The complete diagnosis and reflection session can be presented in 6 stages as shown: (1) Initialize the patient agent with a profile generated through the D$^4$ dataset, and embed the ground truth diagnosis result in the patient agent. The diagnosis results of the patient agent remain unseen to the patient agent and psychiatrist agent until the predicted diagnosis result is generated. (2) The psychiatrist and the patient agent begin the conversation. (3) The supervisor plugin updates the diagnosis tracking list, provides next-turn instruction based on the dialogue history and diagnosis tracking list, and stimulates conversation of the next round. The step (2) and (3) repeat until the conversation finish. (4) After the conversation, the psychiatrist agent generates the electronic medical records and the diagnosis result. The supervisor plugin compares the ground truth diagnosis result and the diagnosis result generated by the psychiatrist agent, and reflects the skills if the generated results are not correct. (5) The supervisor plugin updates the summarized skills in the memory module of the psychiatrist agent, and (6) The psychiatrist agent calls the next patient and repeats the whole procedure.
To improve dialogue quality and diagnosis accuracy, each agent contains a chat module, a memory module, and a retrieval module. Psychiatrist's memory is a newly proposed tertiary memory framework. Each module will be introduced in detail.

\subsection{Chat Module}
The chat module takes the profile of the agent, relevant memory retrieved, and dialogue history as input, and generates the utterance that the agent going to say. Since the psychiatrist always plays a proactive role during the diagnosis conversation, we take advantage of the supervisor plugin of the psychiatrist agent to control the dialogue process. The dialogue generation function used in step (2) of Figure~\ref{fig:overview} can be presented in Equation~\ref{eq:dialogue generation}. The $\mathrm{PPT}_d$, $\mathrm{PRF}$ indicates the constant prompt and personal profile utilized for dialogue generation, respectively, the $\mathrm{mem}$ suggests the retrieved memory, $\mathrm{ins}$ means the instruction generated by supervisor plugin, only provided if the profile is the psychiatrist agent, and the $\mathrm{utt}_{:i-1}$ is the dialogue history until now. 

\begin{equation}
    \mathrm{utt}_{i} = \mathrm{LLM}(\mathrm{PPT}_d, \mathrm{PRF}, \mathrm{mem}, [\mathrm{ins}],\mathrm{utt}_{:i-1})
    \label{eq:dialogue generation}
\end{equation}

The diagnosis function is attached in the chat module. The diagnosis function takes the whole dialogue history and relevant memory retrieved as input, and generates the detailed diagnosed symptoms and the predicted depression risk and suicide risk. The diagnosis function can be presented in Equation~\ref{eq:diagnosis}. The $\mathrm{PPT}_{diag}$ indicates the constant prompt utilized for diagnosis result generation. Since the response from LLM is always unstable, it results in bias during the experiments. Therefore, we apply a voting diagnosis method. To be specific, we let the LLM generate $k$ samples of diagnosis response ($k=5$ in our experiments). After that, we map the depression risk and suicide risk to integer: control(0), mild(1), moderate(2), and severe(3) as shown in Equation~\ref{eq:vote}. We get the most voted risk level in most cases. If tie, we choose the rounded average of all votes as the final decision. 

\begin{equation}
    \mathrm{diag}_i = \mathrm{LLM}(\mathrm{PPT}_{diag}, \mathrm{PRF}, \mathrm{mem}, \mathrm{utt}_{:})
    \label{eq:diagnosis}
\end{equation}

\begin{equation}
\mathrm{diag}=\begin{cases}
      \mathrm{mode}(\{\mathrm{diag}_i\}_k), & \text{if only one mode exists} \\ 
      \mathrm{average}(\{\mathrm{diag}_i\}_k), & \text{otherwise}
    \end{cases}
\label{eq:vote}
\end{equation}

After the conversation, the chat module summarizes the dialogue history as EMR, including all the symptoms and their severity mentioned in the dialogue. The function is presented in Equation~\ref{eq:EMR generation}, the $\mathrm{PPT}_{emr}$ represents the constant prompt utilized for concluding EMR.

\begin{equation}
    \mathrm{EMR} = \mathrm{LLM}(\mathrm{PPT}_{emr}, \mathrm{utt}_{:})
    \label{eq:EMR generation}
\end{equation}

\subsection{Tertiary Memory Mechanism}

Memory module is provided as an external storage space to overcome the limitation of input length of LLMs~\cite{zhang2024surveymemorymechanismlarge}. The previous memory structure~\cite{Park2023GenerativeAgents} regards memory nodes as equal nodes, which is not appropriate in more complex scenarios. Although some works have proposed multi-layer memory structures~\cite{qian2024chatdevcommunicativeagentssoftware,li2023tradinggptmultiagentlayeredmemory,sumers2023cognitive, maharana2024evaluatinglongtermconversationalmemory}, they are still underexplored, as they are based merely on the long-short memory concept without a specific division of roles across different memory layers. To fully simulate the process of psychiatrist training in a real-life scenario, we design a three-level hierarchical memory structure consisting of \textit{Conversation Records, Electronic Medical Records, and Diagnostic Skills}, 
which mimics the exact training process of a psychiatrist, where useful information is extracted into EMR during a patient counseling session, and as their experiences with more patients accumulate, their clinical and diagnostic skills are elevated.  

\begin{itemize}
    \item \textbf{Conversation Records} represent the pure transcripts of the diagnosis conversation between patient agents and psychiatrist agents. It helps the psychiatrist agent to conclude the electronic medical records of the patient agents. 
    \item \textbf{Electronic Medical Records (EMR)} indicate the summary of the patients' portrait, chief complaint, and symptom list. EMR can be leveraged as a summary of specific diagnosis sessions, which can be utilized by psychiatrists to search for similar cases and summarized skills.
    \item\textbf{Diagnostic Skills} play as the optimizer in training psychiatrist agents. Summarized Skills take the dialogue history, diagnosis result generated by the psychiatrist agent, and the ground truth diagnosis result given by a professional psychiatrist in real life in D$^4$ as input, while output summarized skills mainly focus on what kind of symptoms are overestimated or underestimated during diagnosis procedure, guide the LLM-powered psychiatrist agent towards professional psychiatrists. Summarized skills help the psychiatrist agent to imitate the real-world distribution.
\end{itemize}

\begin{figure}[!ht]
    \centering
    \includegraphics[width=1\linewidth]{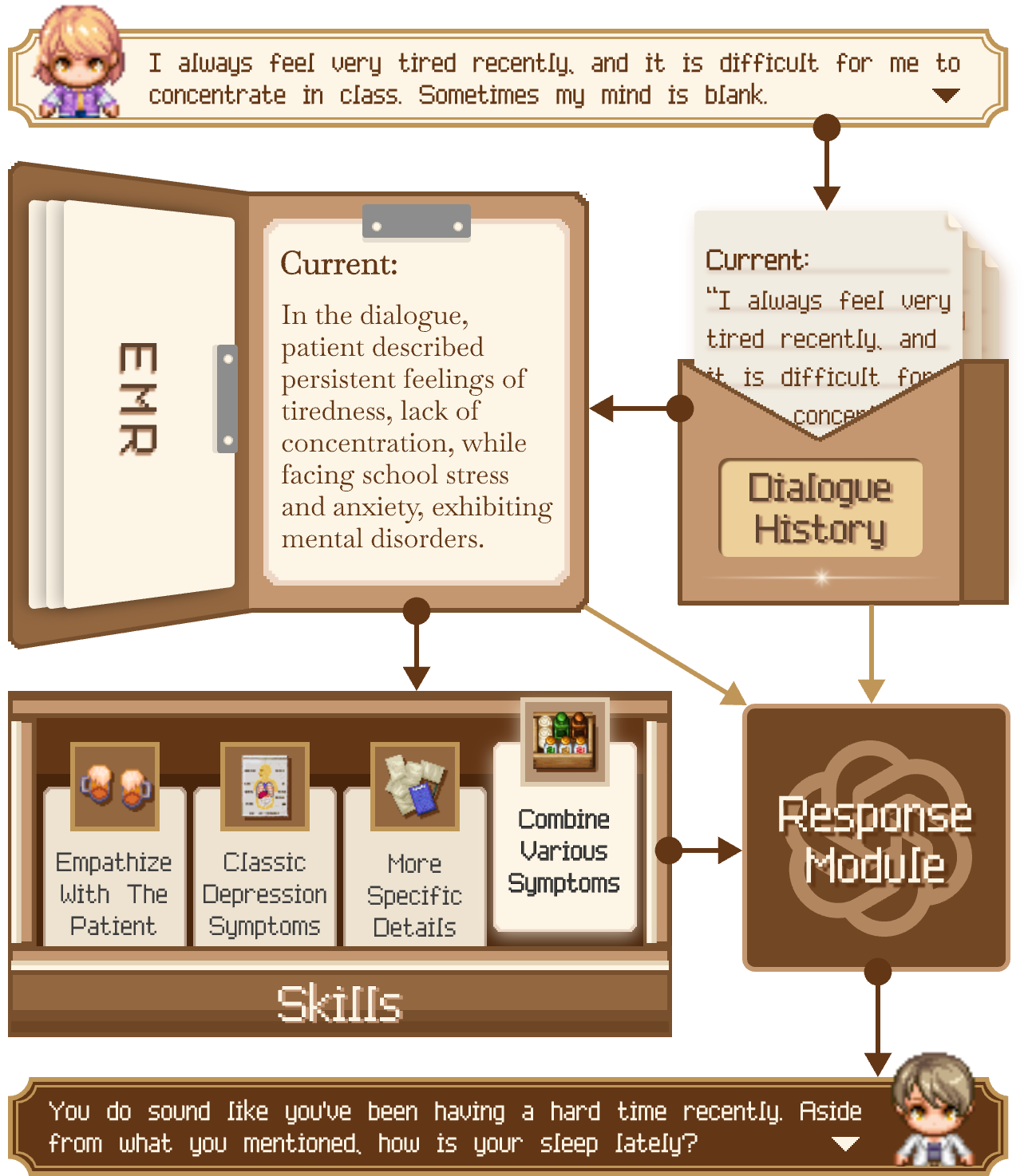}
    \caption{\textbf{The Tertiary Memory Structure of \system.} The utterance of the diagnosis conversation will be stored in \textbf{Dialogue History}. The whole dialogue history in the session will be summarized into electronic medical records (\textbf{EMR}). \textbf{Skills} are generated by the supervisor plugin. All the memory will contribute to the dialogue generation.}
    \label{fig:memory structure}
\end{figure}

As illustrated in Figure~\ref{fig:memory structure}, these three tiers of memory mechanisms constitute a progression from the most fundamental and granular level, through abstraction, to the most condensed, compressed, and consequently, more enduring form. Each subsequent layer represents a refinement process where information is distilled from its original detailed state into a more concise and long-lasting representation. This hierarchical organization enables the psychiatrist system to not only store vast amounts of data efficiently but also retrieve it effectively over time, as the most essential elements are preserved in the most refined and enduring form.

\subsection{Supervisor Plugin of the Psychiatrist Agent}
The supervisor plugin tracks the symptoms mentioned by the patient agent and maintains a detailed symptom list, where each symptom is assigned one of three statuses: unknown, true, or false. Meanwhile, the supervisor plugin updates the list based on the dialogue history, and also provides the question list psychiatrist agents need to ask in the next round, avoiding repeatedly asking similar questions but focusing on those unknown symptoms, which enhances the effectiveness of the diagnosis conversation. The supervisor plugin also controls the dialogue stage change as we define three stages of the diagnosis dialogue:
\begin{itemize}
    \item \textbf{Start Stage}: The start stage aims to start the diagnosis conversation session, mainly focusing on asking the chief complaints of the patient agents and building a healthy therapeutic alliance~\cite{Horvath2011AllianceII} with them. 
    \item \textbf{Exploring Stage}: The exploring stage aims to collect detailed information of the patient agents, and tries to find out the most effective way to complete the diagnosis session.
    \item \textbf{End Stage}: The end stage aims to finish the dialogue, summarizes the information collected, and gives some advice to the patient agents based on the dialogue history. 
\end{itemize}

The instruction generation function of the supervisor plugin can be presented in Equation~\ref{eq:instruction generation}. The $\mathrm{PPT}_ins$ implies the constant prompt utilized in the instruction generation.

\begin{equation}
    \mathrm{ins}_{i}, \mathrm{stat}_i = \mathrm{LLM}(\mathrm{PPT}_c, \mathrm{PRF}, \mathrm{stat}_{i-1}, \mathrm{utt}_{:i-1})
    \label{eq:instruction generation}
\end{equation}

The supervisor plugin also acts as a teacher to generate skills based on the ground truth diagnosis result and the generated diagnosis result based on the dialogue history. The skill generation function is represented in Equation \ref{eq:skill generation}. $\mathrm{PPT}_s$ implies the constant prompt utilized for reflecting skills, $\hat{\mathrm{diag}}$ means the ground truth diagnosis result provided by the supervisor, based on annotation on D$^4$ dataset.

\begin{equation}
    \mathrm{Skills} = \mathrm{LLM}(\mathrm{PPT}_s, \mathrm{PRF}, \mathrm{utt}_{:}, \hat{\mathrm{diag}}, \mathrm{diag})
    \label{eq:skill generation}
\end{equation}

\begin{table*}[h!]
    \centering
    \begin{tabular}{cccccccc}
        \toprule
        \multirow{2.5}{*}{\textbf{Setting}} & \multirow{2.5}{*}{\textbf{Memory}} & \multicolumn{3}{c}{\textbf{Original Dialogues}} & \multicolumn{3}{c}{\textbf{Simulated Dialogues}} \\
        \cmidrule(rl){3-5}\cmidrule(rl){6-8}
        &            & Dep. & Su. & \textbf{Overall} & Dep. & Su. & \textbf{Overall} \\
        \midrule
        \multirow{2}{*}{\makecell[c]{\textbf{Quiz}\\(Train)}}  & w/o &  41.0    &   49.8  & \cellcolor{gray!30}45.4     & 21.8  &    23.4  & \cellcolor{gray!30}22.6     \\
                                        & w/ & 48.2(+7.2)  & 51.4(+1.6)  & \cellcolor{gray!30}49.8(+4.4)     & 27.6(+5.8)  & 26.4(+3.0)  & \cellcolor{gray!30}27.0(+4.4)     \\
        \midrule
        \multirow{2}{*}{\makecell[c]{\textbf{Exam}\\(Test)}}   & w/o  &  28.0    &   26.0   & \cellcolor{gray!30}27.0       &  16.4    &    12.0  & \cellcolor{gray!30}14.2       \\
                                        & w/ & 32.4 (+4.4)  & 27.0 (+1.0)  & \cellcolor{gray!30}29.7(+2.7)     & 23.2(+6.8)  & 13.6(+1.6)  & \cellcolor{gray!30}18.4(+4.2)     \\
        \bottomrule
    \end{tabular}
    \caption{\textbf{The main experiment results on depression diagnosis.} Dep. : the accuracy of depression risk classification. Su. : the accuracy of suicide risk classification.}
    \label{tab:main}
\end{table*}

\subsection{Retrieval Module}
The retrieval module aims to search for helpful electronic medical records and summarized skills, to better generate responses and diagnosis results. The retrieval module is also responsible for the reassignment of the memory nodes, based on the diagnosis result. We modify the classical retrieve score calculation function~\cite{Park2023GenerativeAgents}. However, since our system does not simulate the time step of the diagnosis, we do not apply the calculation of recency in our experiment. Therefore, we mainly calculate relevance score and importance score. The relevance score is calculated as $\mathrm{rel} = \mathrm{NORM}(\mathrm{E}(\mathrm{mem})\cdot \mathrm{E}(\mathrm{query}))$. $\mathrm{NORM}$ indicates min-max scaling method for normalization. $\mathrm{E}$ represents the embedding model, and $\mathrm{query}$ suggests the last utterance during the diagnosis conversation, and the concluded EMR of the dialogue during the diagnosis. The importance score $\mathrm{imp}$ is set to 5 when the memory node is generated, and updates based on the diagnosis results, which suggests the effectiveness of the according memory. The importance score also applies min-max scaling normalization during calculation. Specifically, if the diagnosis result matches the ground truth, the importance score will be increased by one; if it does not, the score will be decreased by one. The final score is calculated as the weighted sum of relevance score $\mathrm{score} = \alpha_1 \mathrm{rel} + \alpha_2 \mathrm{imp}$, while $\alpha_s$ are set to 1 in our experiments. Meanwhile, our observations during the experiments indicate that if we utilize the mere top-$k$ retrieve method as prior work~\cite{Park2023GenerativeAgents} utilized, the memory node retrieved does not vary much with the experiment going, since the top-$k$ memory nodes are fixed. Therefore, we apply a sampling memory retrieval method in the retrieval module. To be specific, we normalize the retrieval score into a probability distribution, which can be seen in Equation~\ref{eq:probability}. After normalizing the score, the retrieval module selects $k$ memory nodes based on the probability distribution. $k$ is set to 10 in our experiment setting. 

\begin{equation}
    p_i = \frac{\mathrm{socre}_i}{\sum_j \mathrm{score}_j}
    \label{eq:probability}
\end{equation}

\section{Experiments}
\label{sec:exp}

To investigate whether the psychiatrist can improve his diagnostic skills, our core experiments lie on whether an \system~ elevates dialogue quality and depression diagnosis accuracy. The general settings, including training strategy and metrics, are introduced below.

\paragraph{Settings}
Since \system's architecture has a supervisor plugin to conduct reflection, we follow the same experiment setting as previous works~\cite{shinn2023reflexion, renze2024selfreflectionllmagentseffects} in the training stage. The psychiatrist agent generates the diagnosis result through Equation~\ref{eq:vote}.

The experiments involve two distinct settings: Quiz and Exam, evaluate on either train set or test set detailed in \S\ref{train/test}. In the \textbf{Quiz} setting, the psychiatrist agent works with the \textit{train set}. After each conversation, it generates a diagnosis result. If the result is incorrect, the agent receives feedback from the supervisor to refine its diagnostic skills and attempts the diagnosis again. This process tests the self-refinement capability of the LLMs, allowing the agent to improve its performance over time, even when applied to unseen patients.

In the \textbf{Exam} setting, the psychiatrist agent is evaluated on the \textit{test set}. Here, the agent generates a diagnosis result based on the knowledge and memory it accumulates from the Quiz setting. Unlike the Quiz, the agent only has one opportunity to provide a diagnosis without a second attempt if the initial result is incorrect. This setting is designed to assess the agent's generalization ability, reflecting the real-world scenario where a diagnosis is made without the chance for revision.

We use \texttt{gpt-3.5-turbo-0125} \cite{chatgpt} for the content generation and \texttt{text-embedding-ada-002} to get embeddings of the content.

\paragraph{Metrics} Two accuracy aspects are involved:
depression risk prediction accuracy and suicide risk prediction accuracy, we also report their average score as overall accuracy. For each patient agent, the risk of depression and the risk of suicide are classified into four categories: control, mild, moderate, or severe.

\paragraph{Results: Does AMC work?}
To extensively validate whether AMC works, we conduct the main experiments under two scenarios:
\begin{enumerate}
    \item \textbf{Original Dialogue (OD)}, \system~takes the original dialogue history in D$^4$ as the conversation between the psychiatrist agent and the patient agent, in order to eliminate the bias of role-playing. The psychiatrist agent conducts the diagnosis directly based on this dialogue.
    \item \textbf{Simulated Dialogue (SD)}, \system~generates simulated diagnosis dialogues between the psychiatrist agent and the patient agent, with the psychiatrist agent performing the diagnosis based on this newly generated dialogue. 
\end{enumerate} 

For each scenario, we construct experiments on two settings: 1) Deactivate the memory module of the psychiatrist agent, evaluating the baseline capability of original LLMs, denoted as \textit{w/o} memory. 2) Activate the memory module and retrieve 10 memory nodes from EMR and skills, denoted as \textit{w/} memory.

We present the results in Table~\ref{tab:main}, which demonstrate the efficacy of the memory module and reflection function on almost all settings, with stable improvements. Across original and simulated dialogues, the largest performance gain from memory and reflection is achieved on \textit{depression diagnosis accuracy}, where even in the most difficult setting (test cases in simulated dialogues), an increase of $6.8\%$ is observed. However, the performance of experiments on simulated dialogues has a significant decrease, despite the improvement of memory structure. The experiment result indicates the limited capability of LLMs to describe the symptoms precisely based on patient profiles. We attach the case study in the appendix, which can further explore the role-play ability of LLMs in diagnostic scenarios.

\textit{\textbf{Take Home Message: } Memory module and reflection function significantly enhance depression diagnosis accuracy, especially in challenging scenarios, but LLMs struggle with precisely simulating symptoms in diagnostic dialogues, highlighting role-play limitations.}

\paragraph{Effects of different memories}

\begin{figure}[!ht]
    \centering
    \includegraphics[width=1\linewidth]{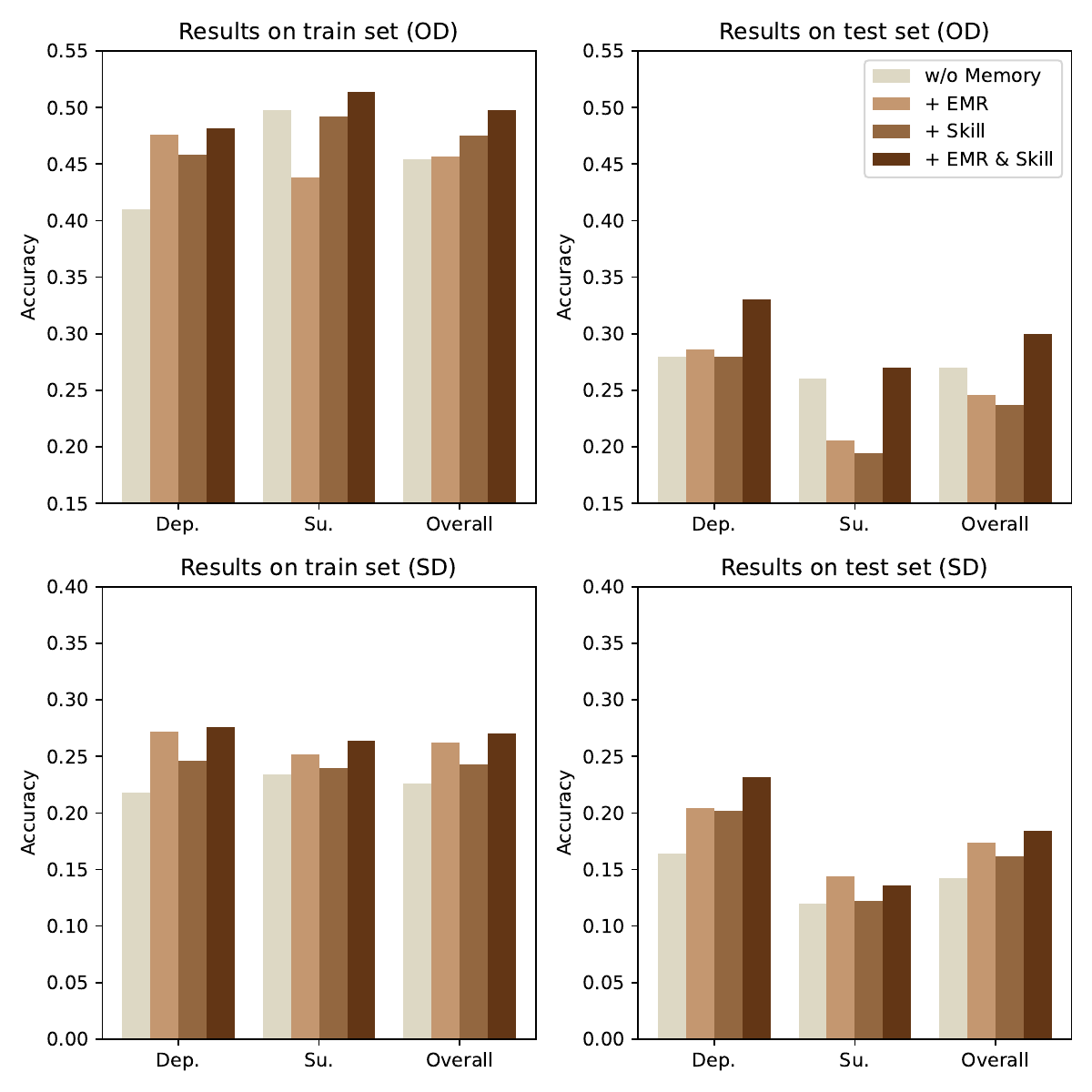}
    \caption{\textbf{The Results of Ablation Study on Memory Layers.} The first row indicates the results based on the original dialogue history setting, while the second row implies the results based on the simulated dialogue setting. At the same time, the first column suggests the results on the train set, while the second column indicates the results of the test.}
    \label{fig:exp}
\end{figure}

To investigate the impact of different memory types on depression diagnosis, we conduct an experiment by manipulating the activation levels of memories: 1) Deactivating the memory module, 2) Activating the memory module and retrieving memories from electronic medical records (EMR), 3) Retrieving memories based on diagnostic skills, and 4) Retrieving memories from both EMR and diagnostic skills. The results of the experiment are depicted in Figure~\ref{fig:exp}. 

From the bar charts, it is evident that in OD settings, diagnostic skills are more effective, likely due to their role in aligning LLMs. Additionally, EMR appears to be more beneficial when dialogues are simulated (SD). This is because the simulated dialogues lack the precision of original diagnosis dialogues, and EMR can assist the psychiatrist agent by retrieving similar patient cases for a more accurate diagnosis. Furthermore, the results show that leveraging both EMR and diagnostic skills yields better and more stable performance than relying on the original LLMs alone.

\textit{\textbf{Take Home Message: } Combining multi-level memory enhances depression diagnosis, ensuring stable and improved performance across settings.}

\paragraph{Effects of supervisor plugin}

To validate the function of the supervisor plugin, we conduct an ablation experiment by disabling the guidance reflection and generation features. The results are presented in Table~\ref{tab:plugin}. The experiment demonstrated that the supervisor plugin improves the accuracy of risk prediction. 

\textit{\textbf{Take Home Message: } The supervisor plugin enhances targeted risk predictions.}

\begin{table}[htbp]
    \centering
    \small
    \setlength{\tabcolsep}{3pt}
    \begin{tabular}{ccccc}
        \toprule
        \textbf{Setting} & \textbf{Plugin} & Dep. & Su.  & \textbf{Overall} \\
        \midrule
        \multirow{2}{*}{\makecell[c]{\textbf{Quiz}\\(Train)}} & w/o & 24.0   & 25.2  &    \cellcolor{gray!30}  24.6   \\
                                        & w/  & 27.6(+3.6)  & 26.4(+1.2) & \cellcolor{gray!30} 27.0 (+2.4)       \\
        \midrule
        \multirow{2}{*}{\makecell[c]{\textbf{Exam}\\(Test)}}  & w/o &  22.0  & 12.0 & \cellcolor{gray!30}17.0      \\
                                        & w/  & 23.2(+1.2) &  13.6(+1.6)    &  \cellcolor{gray!30}18.4(+1.4)          \\
        \bottomrule
    \end{tabular}
    \caption{\textbf{Effects of Supervisor Plugin in Reflection and Providing Feedback.}}
    \label{tab:plugin}
\end{table}

\section{Related Work}
\paragraph{Conversational Agent Simulation System}
Conversational agents (CAs) are agents that can interact with users through natural language~\cite{Allouch2021ConversationalAG}. With the development of large language models, more and more CAs are constructed based on LLMs~\cite{Park2023GenerativeAgents,  chen2023llmempoweredchatbotspsychiatristpatient, shao2023characterllmtrainableagentroleplaying, li2023chatdoctormedicalchatmodel, han2024ibsendirectoractoragentcollaboration, 10.1145/3586182.3615796, jörke2024supportingphysicalactivitybehavior, yang2024simschatcustomisablepersonadrivenroleplaying} because of its autonomy, reactivity, pro-activeness and social ability~\cite{xi2023risepotentiallargelanguage}. Depending on whether controlling the conversation process, CAs can be divided into proactive agents~\cite{Liao2023ProactiveCA, Deng2023PromptingAE, Zhang2024AskbeforePlanPL, Deng2024TowardsHP, deng-etal-2023-knowledge} and inactive agents. In our system \system, psychiatrist agents play the role of collecting information, deciding the dialogue process, and conducting diagnosis, while patient agents need to provide information based on their profiles and the questions.

\paragraph{Mental Health Diagnosis Chatbot}
According to previous research, mental health diagnosis based on dialogue can be more favorable and effective than self-rating scales~\cite{Vaidyam2019ChatbotsAC, Abdalrazaq2019AnOO}, and a better way to elicit honest self-disclosure about personal experiences and emotions~\cite{Kawasaki2020AssessingUM, Duvvuri2022PredictingDS}. Based on whether modifying the weights of language models, the chatbot can be divided into fine-tuning method~\cite{yao-etal-2022-d4, gu2024enhancingdepressiondiagnosisorientedchatpsychological,Lan2024TowardsRA, Ren2024WundtGPTSL} and prompting method~\cite{chen2023llmempoweredchatbotspsychiatristpatient, Tao2023ClassifyingAA, Li2024LeveragingLL, Ferrario2024TheRO, Kumar2024MentalHC, Wang2023ChainofthoughtPF, seo2024diagescdialoguesynthesisintegrating}. Although the fine-tuning method has achieved high diagnosis accuracy and a standardized information-collecting process, the prompting method shows its potential to stimulate the capability of LLMs without modifying the weights, which could be costly as the scale of the language model increases.

\section{Limitation and Future Work}
Our research has limitations that can be further improved in future work.

While the Agent Mental Clinic (AMC) system has shown promising potential in enhancing depression diagnosis through simulated dialogues, some areas warrant further development. The system’s effectiveness is currently shaped by its reliance on the D4 dataset, which may limit its applicability in diverse cultural and linguistic contexts. Additionally, the accuracy of role-playing within the system reflects the inherent challenges of current large language models (LLMs) in fully capturing the details of depressive symptoms, which can sometimes result in less precise simulations. Although the memory retrieval module has seen improvements, it still encounters difficulties fully encompassing mental health diagnoses' complexities. Moreover, the absence of real-time expert feedback may affect the system's ability to adapt to particularly complex cases.

Future efforts would focus on broadening the AMC system's adaptability to various cultural settings, further refining the role-playing capabilities of LLMs, and advancing the sophistication of memory retrieval methods. Integrating real-time expert feedback and extending the system's application to additional mental health conditions could enhance its accuracy, cultural sensitivity, and overall comprehensiveness in diagnostic processes.

\section{Conclusion}

In this paper, we propose a novel conversational agents simulation system named \system, which is designed for depression diagnosis conversations. We construct the psychiatrist agents with tertiary memory structure, and construct supervisor plugin to perform better quality long-term depression diagnosis dialogue. Our experiments demonstrate the better performance of our \system, which could be utilized for simulating the depression diagnosis dialogue in real-life scenarios and getting the diagnosis results based on the conversation history. We conduct the experiments on the original D$^4$ history and generate simulated dialogues, both indicate the effectiveness of the memory structure and the supervisor plugin. In future work, we aim to expand the application of \system~to other domains of mental health and explore the incorporation of additional expert feedback mechanisms to further enhance its diagnostic precision. We believe \system~can be presented as an effective automation in depression diagnosis scenario. 

\section{Acknowledgements}

We are grateful for the accessible OpenAI API key and the visualization provided by Shanda Group. We also appreciate the image materials provided by RPG Maker MZ\footnote{\copyright Gotcha Gotcha Games Inc./YOJI OJIMA 2020}.



\bigskip
\bibliography{aaai25}

\bigskip

\appendix

\section{Case Study}
As we mentioned in the \S4, despite the improvement of memory structure, the performance of experiments on simulated dialogues has a decrease compared with original dialogues, possibly due to the limited role-play capability of LLMs. We will present some cases to illustrate these limitations. 

\subsection{Hallucination}
In the simulated dialogue experiments, we found that when engaging in role-playing, LLMs tend to make unfounded elaborations based on previously mentioned symptoms or events. In the example presented in Table~\ref{tab:hallucination}, the patient only mentioned a past tendency toward self-harm in the original dialogue in D$^4$, which was promptly stopped by friends. However, in the simulated dialogue, the patient agent exaggerated the description on this by stating "It doesn't happen very often, about once a month or so".

\begin{table}[!ht]
    \centering
    \begin{tabular}{m{0.3\linewidth}m{0.6\linewidth}}
    \toprule
         Original Dialogue & I occasionally \textcolor{orange}{have urges to self-harm}, but my friends \textcolor{orange}{have stopped me in time}. \\ 
    \midrule
    \midrule
         Simulated Question & Could you tell me more about how often and in what ways you self-harm? \\
    \midrule
         Simulated Response & It doesn't happen very often, about \textcolor{orange}{once a month or so}. \\
    \bottomrule
    \end{tabular}
    \caption{\textbf{Example of Hallucination}. The content is translated from Chinese utterances.}
    \label{tab:hallucination}
\end{table}

\subsection{Repetition}
We also observed that when generating dialogues, large models tend to replicate the format and content of previous exchanges, leading to repetitive phrasing or responses. This causes the current phase of the conversation to mirror earlier ones, and subsequent dialogues to mirror the current phase, resulting in the entire conversation using similar structures and getting stuck on certain topics. The example is represented in Table~\ref{tab:repetition}.

\begin{table}[!ht]
    \centering
    \begin{tabular}{m{2cm}m{6cm}}
    \toprule
         Psychiatrist Agent & Mr. Lin, \textcolor{blue}{thank you for sharing} about the decrease in your social activities. Besides that, \textcolor{blue}{have you noticed} any mood swings or feelings of depression recently? Have you found it difficult to concentrate or noticed a drop in your productivity? \textcolor{blue}{These changes are important for me to better understand and assess your mental state}. \\
    \midrule
         Patient Agent & \textcolor{orange}{Recently}, I've noticed that I'm not as socially active as I used to be. I'm no longer as enthusiastic about going out or meeting up with friends. This change has been tough for me, and I do feel a bit lonely because of it. \\
    \midrule
         Psychiatrist Agent & Mr. Lin, \textcolor{blue}{thank you for sharing}. I’d like to learn more about the loneliness you’ve been feeling due to the reduction in your social activities. \textcolor{blue}{Have you noticed} a lack of opportunities to connect with friends, or do you feel that this decrease has added extra stress and a sense of isolation? \textcolor{blue}{These details are important for us to better understand your emotional well-being}. \\
    \midrule
         Patient Agent& \textcolor{orange}{Recently}, I've been feeling lonely and stressed. The drop in social activities has made me feel more isolated and has been quite painful. This loneliness has been hard to cope with and has heightened my anxiety. \\
    \bottomrule
    \end{tabular}
    \caption{\textbf{Example of Repetition}. The content is translated from Chinese utterances.}
    \label{tab:repetition}
\end{table}

\subsection{Language Style}
Owing to the lack of original dialogue for a single patient, the data available for LLMs to learn from is also constrained. Consequently, in simulated dialogues, the models often fail to accurately capture the patient's language style and may even use professional medical terminology in their first session, which doesn't align with the patient's identity when describing symptoms, as Table~\ref{tab:language style} represents. This could lead to potential diagnostic inaccuracies.

\begin{table}[!ht]
    \centering
    \begin{tabular}{m{2cm}m{6cm}}
    \toprule
         Patient Agent & Hello, Dr. Bernard. My name is Xiaohong, and I’m \textcolor{orange}{11 years old}, currently a primary school student. Lately, I've noticed a decline in my memory, and I've been experiencing some \textcolor{orange}{psychomotor agitation} and \textcolor{orange}{retardation}. I’m reaching out for help. \\
    \midrule
         Psychiatrist Agent & Hello, Xiaohong. Thank you for sharing this information with me. \\
    \bottomrule
    \end{tabular}
    \caption{\textbf{Example of Language Style}. The content is translated from Chinese utterances.}
    \label{tab:language style}
\end{table}

\section{Data Characteristics of D$^4$}
\begin{figure}[!ht]
    \centering
    \includegraphics[width=\linewidth]{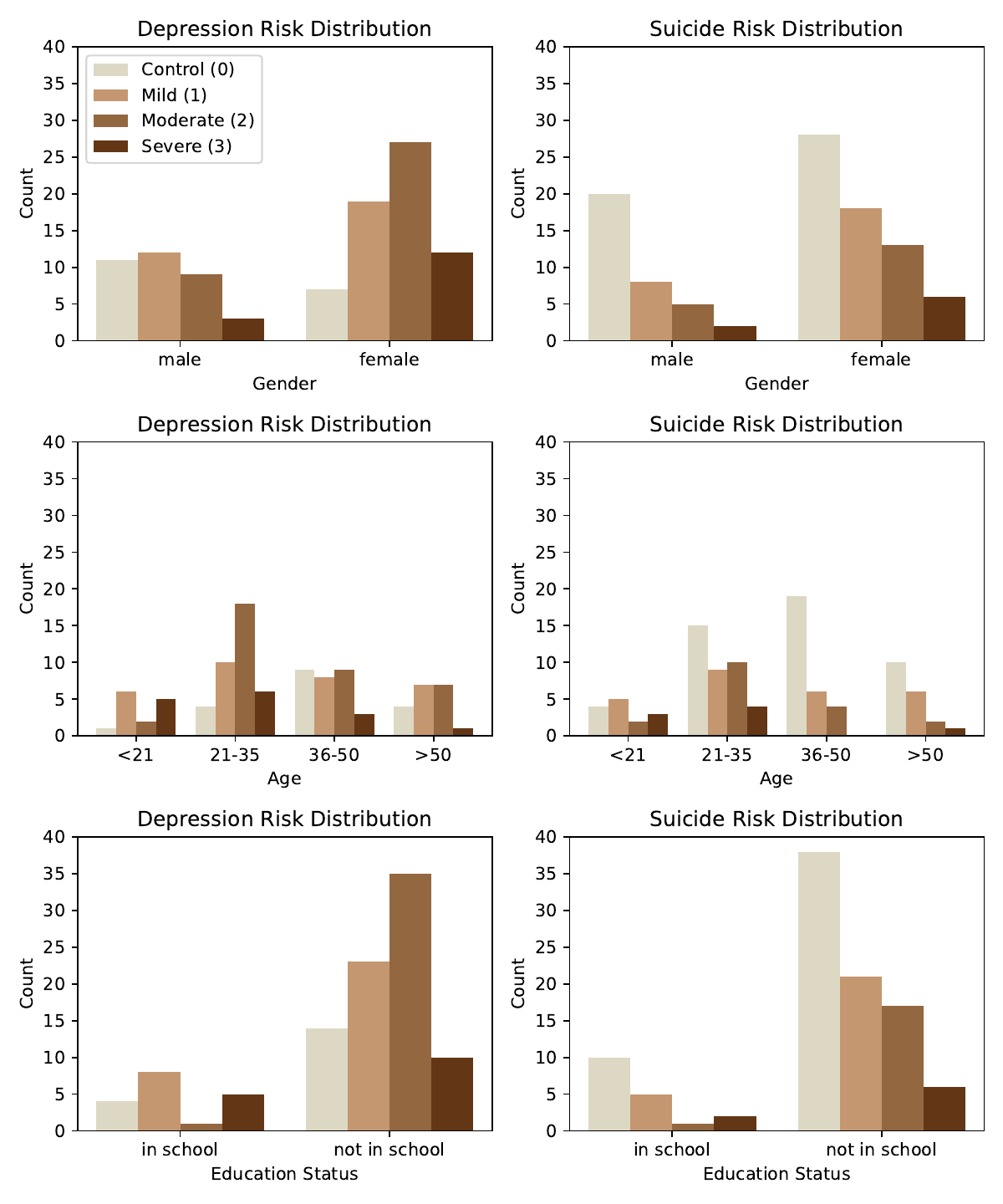}
    \caption{\textbf{The Distribution of Portrait on Train Set.}}
    \label{fig:portrait distribution train}
\end{figure}

\begin{figure}[!ht]
    \centering
    \includegraphics[width=\linewidth]{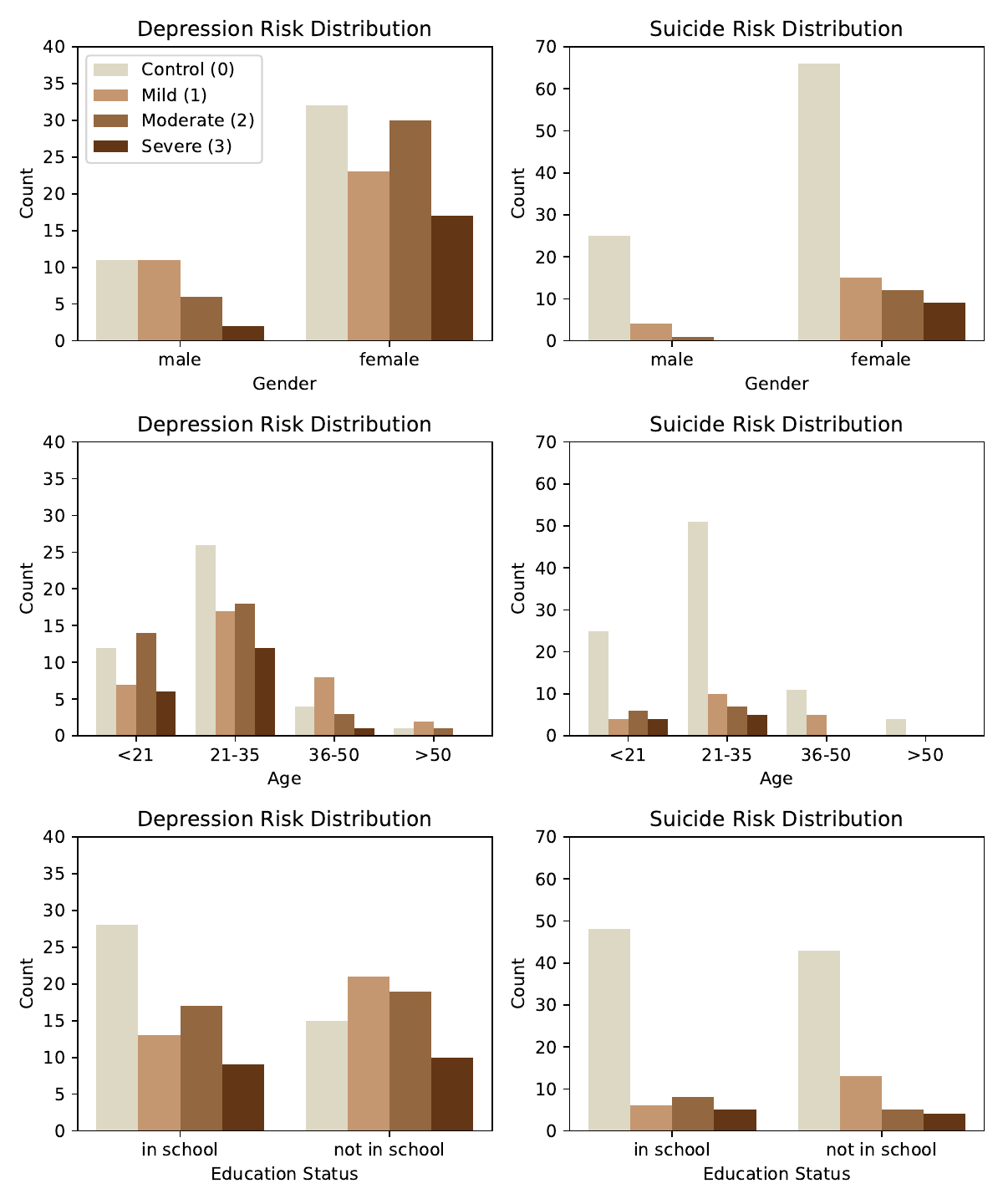}
    \caption{\textbf{The Distribution of Portrait on Test Set.}}
    \label{fig:portrait distribution test}
\end{figure}

D$^4$ is a Chinese depression diagnosis dataset, the overall statistics can be shown as Table~\ref{tab:data characteristic}. The distribution of the User Portrait is presented in Figure~\ref{fig:portrait distribution train} and Figure~\ref{fig:portrait distribution test}.

\begin{table}[htbp]
    \centering
    \small
    \begin{tabular}{cccc}
    \toprule
    \textbf{Category}&\textbf{Total}&\textbf{Patient}&\textbf{Doctor} \\
    \midrule
        Dialogues & 1339 & - & - \\
    \midrule
        Avg. turns & 21.6 & - & - \\
    \midrule
        Workers & 201 & 127 & 74 \\
    \midrule
        Avg. utterances per dialogue & 60.9 & 30.9 & 29.9 \\
    \midrule
        Avg. tokens per dialogue & 877.6 & 381.8 & 495.8 \\
    \midrule
        Avg. tokens per utterance & 14.4 & 12.3 & 16.6 \\
    \bottomrule
    \end{tabular}
    \caption{\textbf{Overall Statistics of D$^4$.}}
    \label{tab:data characteristic}
\end{table}

\section{Ethics and Broader Impact Statement}

    \paragraph{Data Privacy}
    The data used in this paper comes from a previously published dataset, and all evaluation metrics are based on objective computations. Although experts' evaluations could be included in future work, this study currently raises little ethical concerns. Any content related to user privacy has been appropriately processed for disclosure in the paper and supplementary files.

    \paragraph{Intended Use}
    \system~system is designed to conduct simulated depression diagnosis dialogues and assess both the depression risk and suicide risk of potential patients. The system can be subsequently be deployed as interactive system for intern psychiatrist training and preliminary screening for potential patients.

    \paragraph{Potential Misuse}
    Since the \system~system is designed to conduct the preliminary screening, the psychiatrist agent is equipped with diagnosis knowledge collected from ICD-11. Therefore, the psychiatrist agent is not appropriate to provide medical suggestions, which might lead to misinterpretation or misuse of the information provided. Instead, the psychiatrist agent is intended to offer general guidance and support during the initial stages of assessment, while encouraging users to seek professional medical advice from a qualified healthcare provider for any specific concerns or diagnoses. This approach ensures that the system serves as a complementary tool, rather than a substitute for professional medical consultation.

    \paragraph{Environmental Impact}
    Our \system~can be utilized for potential patient who are suffering from depressive mood. Currently, mental health issues, especially depressive disorder, are prevalent, yet a significant number of individuals fail to recognize their own mental health problems promptly or seek timely treatment. The \system~system, constructed as a future platform accessible across three interfaces, aspires to serve as an efficient tool for enhancing psychiatrists' professional expertise, offering the public convenient and highly confidential diagnostic assessments, and improving diagnostic accuracy across the mental health field. Through these contributions, we hope to contribute to the improvement of global mental health.

\end{document}